\documentclass[times,twocolumn,final]{elsarticle}

\usepackage{cag}
\usepackage{framed,multirow}

\usepackage{amssymb}
\usepackage{amsmath}
\usepackage{latexsym}
\usepackage[utf8]{inputenc}
\usepackage[english]{babel}

\newcommand{\norm}[1]{\left\lVert#1\right\rVert}

\usepackage{url}
\usepackage{xcolor}
\definecolor{newcolor}{rgb}{.8,.349,.1}

\usepackage{hyperref}

\usepackage[switch,pagewise]{lineno} 

\usepackage{siunitx}
\usepackage{booktabs}

\journal{Computers \& Graphics}

\begin{document}

\verso{Preprint Submitted for review}

\begin{frontmatter}

\title{Spatially and color consistent environment lighting estimation using deep neural networks for mixed reality}%

\author[1]{Bruno Augusto \snm{Dorta Marques}\corref{cor1}}
\emailauthor{bruno.marques@ufabc.edu.br}{Bruno A. D. Marques}
\author[2]{Esteban Walter \snm{Gonzalez Clua}\fnref{fn1}}
\fntext[fn1]{e-mail: esteban@ic.uff.br}
\author[2]{Anselmo Antunes \snm{Montenegro}\fnref{fn2}}
\fntext[fn2]{e-mail: anselmo@ic.uff.br}
\author[2]{Cristina \snm{Nader Vasconcelos}\fnref{fn3}}
\fntext[fn3]{e-mail: crisnv@ic.uff.br}

\address[1]{CMCC, Universidade Federal do ABC, Avenida dos Estados, 5001, Santo André, SP, Brazil}
\address[2]{Instituto de Computação - Universidade Federal Fluminense, Av. Gal. Milton Tavares de Souza, Niterói, RJ, Brazil}

    


\received{\today}

\begin{abstract}
 The representation of consistent mixed reality (XR) environments requires adequate real and virtual illumination composition in real-time. Estimating the lighting of a real scenario is still a challenge. Due to the ill-posed nature of the problem, classical inverse-rendering techniques tackle the problem for simple lighting setups. However, those assumptions do not satisfy the current state-of-art in computer graphics and XR applications. While many recent works solve the problem using machine learning techniques to estimate the environment light and scene's materials, most of them are limited to geometry or previous knowledge. This paper presents a CNN-based model to estimate complex lighting for mixed reality environments with no previous information about the scene. We model the environment illumination using a set of spherical harmonics (SH) environment lighting, capable of efficiently represent area lighting. We propose a new CNN architecture that inputs an RGB image and recognizes, in real-time, the environment lighting. Unlike previous CNN-based lighting estimation methods, we propose using a highly optimized deep neural network architecture, with a reduced number of parameters, that can learn high complex lighting scenarios from real-world high-dynamic-range (HDR) environment images. We show in the experiments that the CNN architecture can predict the environment lighting with an average mean squared error (MSE) of \num{7.85e-04} when comparing SH lighting coefficients. We validate our model in a variety of mixed reality scenarios. Furthermore, we present qualitative results comparing relights of real-world scenes.
\end{abstract}

\begin{keyword}
\KWD Lighting Estimation, Spherical Harmonics, Deep Learning, Environment Map, Mixed Reality, Augmented Reality
\end{keyword}

\end{frontmatter}


\begin{figure*}
  \includegraphics[width=\textwidth]{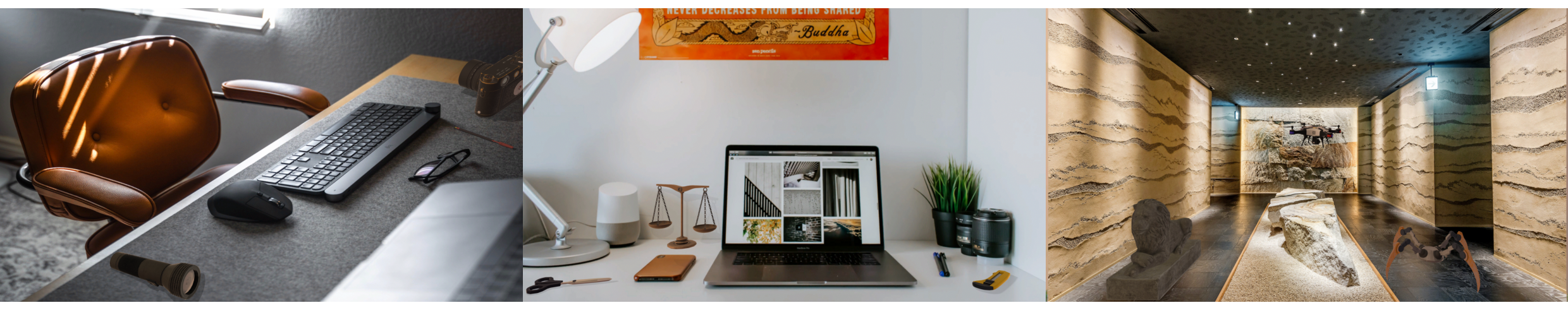}
  \caption{Deep neural network spherical harmonics lighting estimation on mixed reality scenes. Stock photos with virtual objects lit by our lighting estimation model.}
  \label{fig:teaser}
\end{figure*}

\section{Introduction}\label{sec_introduction}

Consistent environment lighting is a crucial component in real-time simulations based on mixed reality applications.  The divergence between real and virtual objects lighting is a significant factor for immersion loss and a perceptual reduced graphical quality \cite{kruijff2010perceptual}. Plausible mixed reality lighting can be accomplished by acquiring the lighting of the real environment and adapting the virtual environment with matching lighting properties \cite{jacobs2006classification}. Most lighting recovery approaches have focused on intrusive tools to measure the environment lighting, requiring great user's effort and scenario preparation. Consequently, these solutions have limited applicability in XR systems based on real-time visualization. An alternative to real-data measurements is to estimate the lighting indirectly through the available environment information. Despite the recent advances in computer vision and inverse rendering \cite{patow2003survey}, estimating the environment lighting without specialized equipment and under strict time constraints remains a challenging problem \cite{ren2016survey}. 
This work aims to recognize the user's environment lighting through a model that learns the scene's inherent characteristics regarding lighting and illumination, therefore estimating an environment lighting capable of generating plausible XR environments.  A challenging aspect of the problem resides in the fact that lighting estimation is an ill-posed problem, yielding no solution or multiple solutions for a given input \cite{ramamoorthi2001signal}. 

We use machine learning techniques and a specialized dataset to overcome the complex aspects of lighting estimation, learning from promptly available information in mixed reality applications: an RGB image of the environment taken from an egocentric point-of-view.

We leverage state-of-the-art lighting estimation methods by predicting the real-world environment lighting using a convolutional neural network that works in the wild without assumptions about the scene's geometry or special measurement devices. Our method does work in a variety of environments, including indoor and outdoor scenes, and does not require any user's intervention in the scene. Our custom-designed CNN architecture learns a latent space representation of the environment lighting, allowing an efficient representation of the scene illumination. This representation is used to estimate the environment lighting encoded in a spherical harmonics basis. We also present a framework to create a mixed-reality-view, an image that mimics the user's egocentric view in an XR environment.

 Figure \ref{fig:teaser} illustrates examples where virtual objects are illuminated by our method. The composition of real and virtual objects can be utilized as a plausible and realistic XR environment.
 
 The main contributions of our work are:

\begin{itemize}
    \item An automatic end-to-end method to estimate the environment lighting in an XR application.
    \item A novel, custom-designed CNN architecture that learns a latent-space representation of environment lighting
    \item A methodology to generate egocentric mixed-reality-views from HDR panoramas. \emph{i. e.}, mixed-reality-view.
    \item A methodology for lighting estimation that works in real-time and does not make restrictive assumptions about the mixed-reality scene neither the application's domain.
\end{itemize}

The lighting estimation model developed in this work can be employed in most XR applications increasing the user's immersion by providing lighting consistency. The applicability of our model is not restricted to mixed reality; other applications also benefit from it, including real-time editing of video and photo with consistent illumination, real-time relighting of pictures, and inverse lighting design \cite{fernandez2012inverse}.

\section{Related Work}\label{sec_related}
Many related works try to solve the lighting estimation task based on different assumptions or strategies. In the following subsections, we group them into categories comparing with our proposed solution. In addition, we highlight the limitations and restrictions of the prior works concerning XR applications when appropriate. 

\subsection{Device-based light probe}
Device-based techniques comprehend methods that make use of a special device that acts as a light probe or directly measures the lighting condition of the scene. In the work of Debevec \cite{debevec2008rendering}, a mirror ball acts as the capturing device, providing the radiance of the environment. Several pictures of the device with different exposure levels are necessary to generate an environment map. Another approach for consistent lighting in XR is acquiring the environment lighting through the image of a wide-angle camera, such as a fish-eye lens camera \cite{walton2018dynamic, kan2013differential, knecht2012reciprocal}. Some other methods focus on designing light probes that can accurately estimate the environment lighting \cite{calian2013shading, aittala2010inverse}. Another approach is to estimate the lighting from single exposure images based on physical objects with different reflectance, such as in DeepLight\cite{LeGendre_2019_CVPR}, that utilizes a capture apparatus that introduce three spheres with different reflectance into the scene. 

Compared to our work, these methods result in accurate environment lighting. However, the usage of such devices to determine the lighting is impractical in many real applications, such as XR environments. Furthermore, those techniques do not work in unprepared environments, restricting their practical usage. Different from the previous works, we choose our approach to be practical and effortless for the end-user.

\subsection{Scene and object geometries}
Some lighting estimation approaches rely on a known geometry of the real scene, or physical object presents in the scene. The assumption of known geometry is a convenient prior for inverse rendering techniques. 

Mandl \emph{et al.} \cite{mandl2017learning} explore known physical objects as light probes to estimate the environment lighting. An object with known geometry is placed in the scene, and a neural network estimates the lighting setting. Weber et al. \cite{weber2018learning} propose a latent space representation of the environment lighting that can be used to estimate the environment lighting of an object with known geometry and reflectance.

Some methods do not assume a previously known scene (or object) geometry but utilize a rough estimation of the depth or surface normals in real-time \cite{whelan2016elasticfusion, meilland20133d, gruber2014efficient, gruber2012real, zhang2016emptying, Maier_2017_ICCV}. These methods utilize depth sensors (RGB-D cameras) to capture scene information. Although there have been advancements of depth sensor technologies, most XR hardware does not provide such capabilities; moreover, those techniques usually require the scan of the complete scene before a reasonable lighting estimation becomes possible. Furthermore, depending on the camera device, environments with high infra-red light incidence, such as the sunlight, may compromise those solutions.

\subsection{Face, eyes and hands light probes}
Lighting estimation is a task that arises from multiple domains. In particular, several works concentrate on lighting estimation for human face images \cite{zollhofer2018state}. Relighting of portraits is a common problem that requires the lighting estimation of a scene. Usually, the 3D morphable model of faces \cite{blanz1999morphable} is utilized to perform an analysis-by-synthesis capable of estimating the environment lighting \cite{shahlaei2015realistic, conde2015efficient, shahlaei2016lighting, egger2018occlusion}. Recently, Sun et al. \cite{sun2019single} presented an alternative methodology for portrait relighting that employs a convolutional neural network to estimates the environment lighting. Those methods rely on images of human faces in portrait poses to estimate the environment lighting. 

The usage of human eyes as light probes is another approach that has been investigated \cite{tsumura2003estimating, nishino2004eyes}. The human eye has a known geometry that is well approximated by a sphere. Furthermore, the high specular reflectance makes the eye a suitable probe for lighting estimation \cite{wang2008separating}. The downside of this approach is the reduced pixel resolution and the limiting factor that a human face should be visible in the image. Those limiting assumptions, especially the latter, are hard to circumvent in a mixed-reality egocentric view perspective.

Finally, the availability of human hands is an assumption made by Marques \emph{et al.} \cite{marques2018shceg, marques2018pls}, where human hands act as implicit light probes for estimating the environment lighting for XR applications. They estimate the lighting through a CNN model that uses images of the user's real hands as input. Although the hands are the main form of interaction in egocentric XR applications, cameras with low field-of-view can limit the applicability of those methods in real applications.

The methods mentioned in this subsection are specific for certain domains where faces, hands, or eyes are visible. These limitations severely restrict the usage of previous methods in general XR applications.

\subsubsection{Outdoor environment}
Lighting estimation for the outdoor environment is another specific domain that several authors explored \cite{Zhang_2019_CVPR,  Hold-Geoffroy_2019_CVPR, lalonde2009outdoor, lalonde2012outdoor}. They tackle the problem by fitting specific parametric outdoor lighting models \cite{perez1993all, habel2008efficient} that take into account sun and skylight properties. However, those methods are out of the scope of our work, considering that our goal is to estimate lighting for arbitrary XR environments (including indoor environments).

\subsection{Surface reflectance and lighting}
Since the environment lighting has a significant influence on the surface reflectance, recovering surface reflectance correlates to lighting estimation tasks. Examples of such correlations are surface reflectance methods that use planned lighting settings under a controlled environment to infer material properties \cite{Deschaintre2018, li2018materials, chen2014reflectance, guarnera2016brdf, ghosh2009estimating}. Another typical context for surface reflectance estimation is to estimate the surface reflectance under unknown environment lighting; in this context, the methods estimate the reflectance and environment lighting simultaneously, using either the geometry of a 3D scanned object \cite{chen2014reflectance, lombardi2015reflectance} or depth sensors \cite{jiddi2016reflectance, Lombardi2016}.  Thus, those methods share the limitations present in the geometry-based lighting estimation methods.

\subsubsection{Environment lighting based on deep learning}
The methods described in this subsection are the closest prior-art that are suitable for XR environments.

Gardner et al. \cite{gardner-sigasia-17} propose a lighting estimation specific to indoor environments; the method uses a CNN to estimate a coarse environment mapping of the scene; the method achieves good results for relights of indoor environments. However, it relies on an intermediate process that makes use of a light classifier to annotate the location of light sources; this process is a possible limiting factor for the applicability of the method in outdoor scenes and non-conventional light sources (indirect light sources, resulting from global illumination effects). 

Song and Funkerhouser \cite{song2019neural} estimate the lighting in a localized point of an indoor scene by decomposing the problem as the subtasks of geometry estimation, warped panorama completion, and LDR to HDR estimation. Decomposing the problem in simpler subtasks helped the learning process, producing higher quality estimations when compared to Gardner et al. \cite{gardner-sigasia-17} work. 

Garon et al \cite{garon2019fast} estimate local lighting of the indoor scene's by combining local and global lighting; this is accomplished by providing both the global image and a local patch of the image to a two-path neural network. The training process of this network requires light probes with depth maps of the scene. To circumvent the laborious process of capturing the light probes and depth maps on real scenes, they make use of synthetic data to generate the required training data. One limitation of Garon et al. lighting prediction is the low color consistency in some cases; they credit the failure cases to their methodology of training the neural network using synthetic data leading to prediction errors in the real test data.

Unlike Garon et al. \cite{garon2019fast} method, we train our CNN on real environment panoramas, and we further propose a color regularization that circumvents their limitations. The usage of environment panoramas also facilitates the process of generating samples for our training dataset.

\begin{figure*}[ht!]
  \centering

  \includegraphics[width=0.9\linewidth]{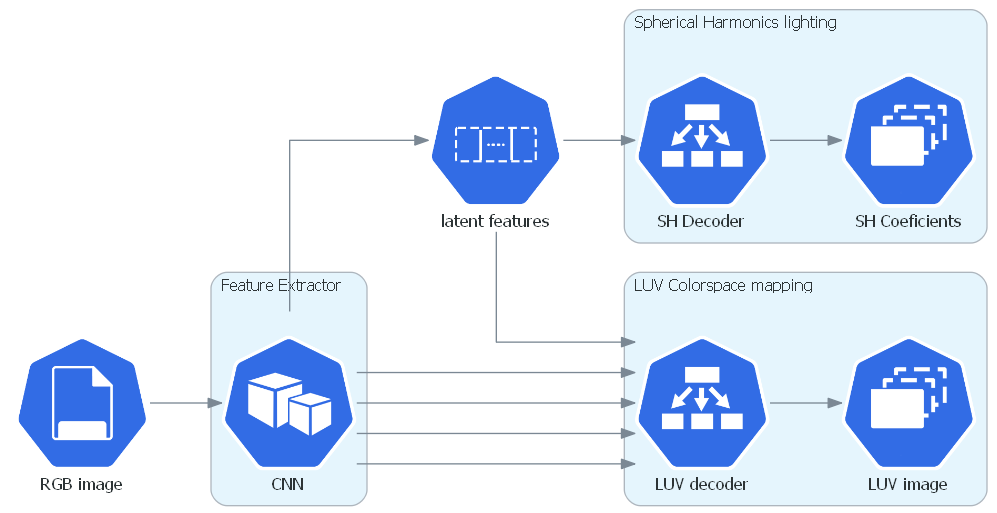}

  \caption{\label{fig:cnn_arch}
           Overview of the lighting estimation CNN architecture. The model's input is an RGB image; the feature extractor extracts the latent features that better describe the input image's environment lighting. The architecture has an SH decoder that describes the environment lighting as a set of spherical harmonics coefficients and an auxiliary LUV decoder that acts as a color regularizer during the training of the CNN.}
\end{figure*}

\section{A CNN method for environment lighting estimation based on spherical harmonics functions}\label{sec_method}
Our goal is to recognize the real-world environment lighting, leveraging this lighting information to virtual environments, allowing more convincing lighting composition for XR experiences. We explore spherical harmonics functions to encode the environment lighting into a compact and expressive representation. This strategy allows representing smooth arbitrary area lighting, not limited to a few point light or directional light
sources \cite{kautz2002fast}.

Our model is based on a convolutional neural network capable of predicting the spherical harmonics environment lighting operating over a single low-dynamic-range (LDR) image. The CNN operates over LDR images, producing plausible environment lighting arrangements with respect to direction, color, and intensities.

To produce realistic lighting scenarios, we use real environment panoramas captured in a great variety of places and lighting settings.  Those panoramas are processed to mimic the images captured by the Head-Mounted-Displays (HMD) in the runtime of XR applications. We name the processed image as mixed-reality-view. The processing of an HDR panorama into mixed-reality-views is described with more details in Section \ref{sec_learn}.

A projection of spherical harmonics functions can describe the lighting settings in a given panorama \cite{Ramamoorthi:2001:ERI}. This operation produces a set of SH coefficients. Our dataset, which is composed of mixed-reality-view and SH coefficients, is employed for training our lighting estimation CNN, as described in Section 5. We demonstrate our lighting estimation model in Section 6 by a set of quantitative and qualitative experiments. Furthermore, we show an application that relights XR environments, and we make a comparison of our method and other state-of-the-art approaches for real-time lighting estimation \cite{gardner-sigasia-17, garon2019fast}.


\subsection{Environment Lighting Estimation}\label{sec_method_lighting}

The environment lighting is the outcome of a complex combination of physical interactions between light sources and surfaces of objects in the scene. There is a variety of representations that can be used to represent the environment lighting, ranging from simplified models (\emph{e.g.} directional light models) to more complex and physically accurate models (\emph{e.g.} resulting radiance from global illumination algorithms such as radiosity and photon mapping \cite{Ritschel2012gi}). In real-time applications, environment lighting is commonly represented by environment maps, which allows image-based lighting (IBL) \cite{debevec2002image} to be implemented in the rendering process. A single environment map can use high-resolution HDR images, in the order of $8192 \times 4096$ pixels. However, a lower sampling rate can produce a good quality environment lighting. A viable approximation of environment maps is the usage of spherical harmonics functions. 
Any signal can be approximated by projecting it onto spherical harmonics basis functions. In particular, low-order functions are sufficient to approximate the environment lighting due to the diffuse nature of the signal \cite{Ramamoorthi:2001:ERI}. We choose to use this approximation to represent the environment lighting in our work. The second-order spherical harmonics functions are capable of representing the diffuse environment lighting in a compact format with only nine coefficients per color channel and allow rendering techniques such as pre-computed radiance transfer functions \cite{Sloan:2002:PRT}. Furthermore, we argue that learning a low-dimensional representation of lighting is a more manageable task than a complex high-dimensional function.

\begin{figure}[h!]
  \centering
  \includegraphics[width=0.9\linewidth]{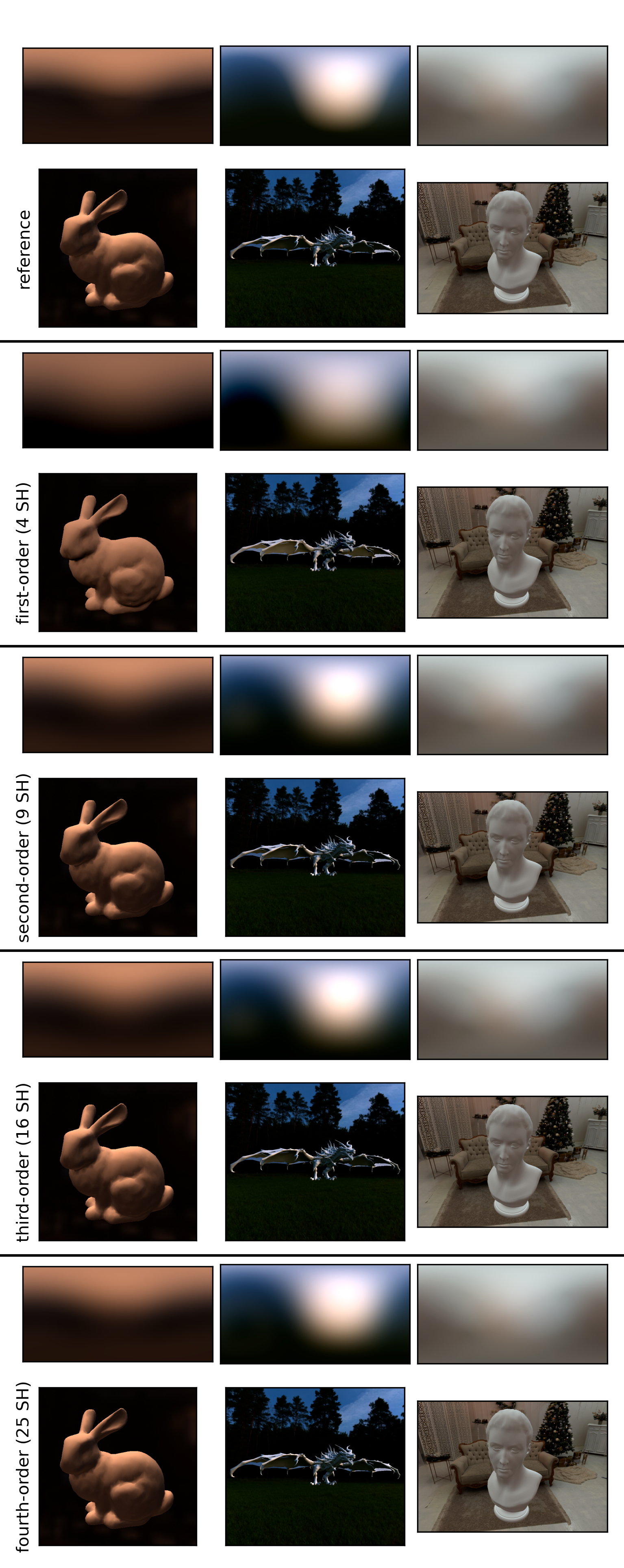}

  \caption{\label{fig:irradianceSH}
           Reference irradiance map (obtained through Monte Carlo importance sampling) and SH approximated irradiance maps with orders ranging from 1 to 4. The renders use the irradiance map as the environment lighting source of the scene.}
\end{figure}

We empirically found that using 9 SH coefficients gives a good compromise between rendering quality and representation size. Figure \ref{fig:irradianceSH} illustrates our findings. We generate and compare environment map representations ranging from first-order SH (4 coefficients) to fourth-order SH (25 coefficients). For the bunny render, the first-order SH approximation has a mean squared error (MSE) of 0.0068 compared to the reference obtained by monte carlo importance sampling. In contrast, the subsequent SH approximations have substantial lower MSE (0.0002, 0.0002, 0.0001). We also rendered a scene to qualitatively compare the rendering quality. While the first-order SH approximation is not sufficient to represent the environment lighting in the scene, the subsequent approximations (orders ranging from 2 to 4) produces a very similar rendering. Similar performance is observed for the dragon render (middle column of Figure \ref{fig:irradianceSH}) with MSE of 0.0084, 0.0004, 0,0004, 0,0001; and the statue (right column) with MSE of 0.0014, 0.0002, 0,0002, 0.0001.

\subsection{Lighting Estimation CNN Architecture}

 The architecture comprises a convolution-based feature extractor and two heads predicting the SH lighting coefficients and a color-space transformation of the input image. 

The recent advances in deep learning allow the development of highly optimized neural networks capable of learning complex tasks with a reduced number of parameters; the SqueezeNet \cite{SqueezeNet}, originally employed for image classification and object detection, is designed to achieve a high level of accuracy while maintaining a low memory footprint. The resulting outcome is a lightweight architecture with comparable accuracy and fewer parameters compared to equivalent bigger neural network architectures. Another significant advantage of this network is the efficiency in the training and inference time of the CNN model. Due to the restrictive time budget of our application (rendering frames in less than 16 ms for a comfortable XR experience), we choose the SqueezeNet as our main convolutional feature extractor. 

Since we are operating on color images, our model's architecture operates in the RGB color channels with an SH decoder that outputs $3\times9$ SH coefficients. An important feature of our model is the color consistency of the lighting estimation. The color consistency consists in predicting plausible lighting estimation in respect to three color channels, without a color channel dominating another one (for example, predicting higher values for the red color while maintaining the green and blue channels with near-zero values). It is important to note that our method is not limited to 2nd order SH representation. Different SH estimations can be done by simply changing the last layer of the SH decoder of our neural network architecture to the appropriate number of SH coefficients. 

Based on previous experiments, we hypothesize that learning only SH coefficients would result in low color-consistency predictions. We employ a second decoder as a color consistency regularizer for our model; this decoder takes the latent-space vector and synthesizes the input image in a different color space (LUV). We hypothesize that the neural network better understands the relationship between the color channels and the lighting in the scene by learning this color-space transformation concurrently to the SH coefficients. The reasoning to use the CIELUV color space in our work is that it is a colorspace with a perceptual lightness component. Still, we believe that any color space with a mapping of perceptual lighting and chromatic components (including CIELAB color space) would result in a similar outcome. Figure \ref{fig:cnn_arch} illustrates an overview of the model's architecture, including the main feature extractor, the SH decoder, and the LUV decoder.  

The SH decoder head is composed of three consecutive fully connected layers with rectified linear activation function and dropout regularizer \cite{srivastava2014dropout}. The fully connected layers have 2048, 1024, and 27 neurons in the hidden layers, respectively. The final layer of the decoder is activated by a Softsign function $ f\left(\bar{y}\right) = \left(\frac{\bar{y}}{|\bar{y}|+1}\right)$. The Softsign function restricts the SH coefficients to $[-1, 1]$ values and have a smoother asymptotic line when compared to the hyperbolic tangent function \cite{lin2018research}.

We model our solution with a regression of the $3 \times 9$ spherical harmonics lighting coefficients and the LUV image. The loss function of our network is determined by the weighted sum of the loss for each decoder head, thus:
\begin{equation}
L = \alpha L_{S} + (1-\alpha) L_{L},
\end{equation}
where $L_{S}$ is the SH decoder loss, $L_{L}$ is the LUV decoder loss and $\alpha$ is the weight scalar.

Since we are modeling the problem with second-degree spherical harmonics functions, we can separate them by three bands. Let $k \in {R,G,B}$ be the index that identifies each lighting channel. The band $0$ with coefficient $C^0_k$ corresponds to the ambient light term, a constant value across all the environment. Band $1$ with coefficients $C^1_k$ correspond to lighting lobes aligned to horizontal, vertical, and depth axis.  The terms of band $2$, $C^2_k$, correspond to the remaining five lighting lobes constituting multiple combinations of directions for each one of the channels. Each coefficient is defined independently for each channel.  
A possible approach is to attribute a weighted scalar to each band in the loss function. Therefore, the loss function per channel results in:
 \begin{equation}
L_i = \sum_{k \in \{R,G,B\}} \alpha E(C^0_k) + \beta E(C^1_k) + \gamma E(C^2_k),
\end{equation}

where $E(c)$ is the mean squared error (MSE).

This formulation for the loss function requires three weighted scalars ($\alpha$, $\beta$, $\gamma$) as training parameters of our model. Since we do not know a priori the optimal value for those scalars, a grid search or a learnable parameters approach is necessary. Instead of evaluating or model predictions directly with this loss function, we choose to employ a differential-rendering approach. The differential-renderer keeps tracking all the operations in the rendering process and calculates the gradient of those operations in the backtracking step of the training process. This approach allows the model to evaluate the predictions and change the model weights based on changes per pixel in the SH environment mapping. We use the same average $E(c)$ metric of the estimated and ground truth environment mapping.

\section{Learning from HDR panoramas}\label{sec_learn}
In this section, we describe the complete pipeline to process the input HDR environment panorama into mixed-reality-views and the corresponding environment lighting.  The mixed-reality-view (MRV) is a low-dynamic-range (LDR) color image similar to a photograph taken from a camera located in the HMD capturing an egocentric view of the user's environment. Spherical harmonics coefficients encode an area light model that represents the environment lighting. Those data are used for training our lighting estimation model. 

Our goal is to generate sufficient data to model arbitrary environmental lighting settings, capturing the implicit casual relationship between the environment appearance and light sources in the scene. We accomplish this from real-world HDR environment panoramas present in the Laval indoor HDR database \cite{gardner-sigasia-17}. Our pipeline generates data with sufficient diversity regarding the user's position and orientation, surface and materials properties, and lighting characteristics. Hence, delivering adequate training data for the lighting estimation model.

The usage of HDR panoramas is fundamental to the lighting estimation process. We use the high-dynamic-range images to generate a Spherical Harmonics representation of the environment that captures all the lighting information of the scene. The ultimate goal of our CNN is to learn the HDR SH representation from conventional LDR images. 

For lighting estimation purposes, it is essential to consider a wide range of light sources with distinctive intensities. Low-Dynamic-Range (LDR) images are widespread and highly available, such as photographs captured by consumer cameras or smartphones. However, LDR photographs can not capture in the same picture the brightest spot of a lamp and the dark details of a shadowing area, for example. 

For our purpose, we use the Laval indoor HDR dataset \cite{gardner-sigasia-17}, a dataset that consists of a large set of high-resolution indoor panoramas captured in High-Dynamic-Range (HDR). The panoramas capture the entire Field Of View (FOV),  with an azimuthal angle of $360$ degrees. There are $2142$ panoramas captured at the resolution of $2048 \times 1024$ pixels. The scenes have a variety of lighting settings ranging from artificial light sources (ceiling, wall, and table lamps) to natural ones (open window, glass door).

\subsection{Mixed-reality-views}

\begin{figure}[h]
  \centering
  \includegraphics[width=0.7\linewidth]{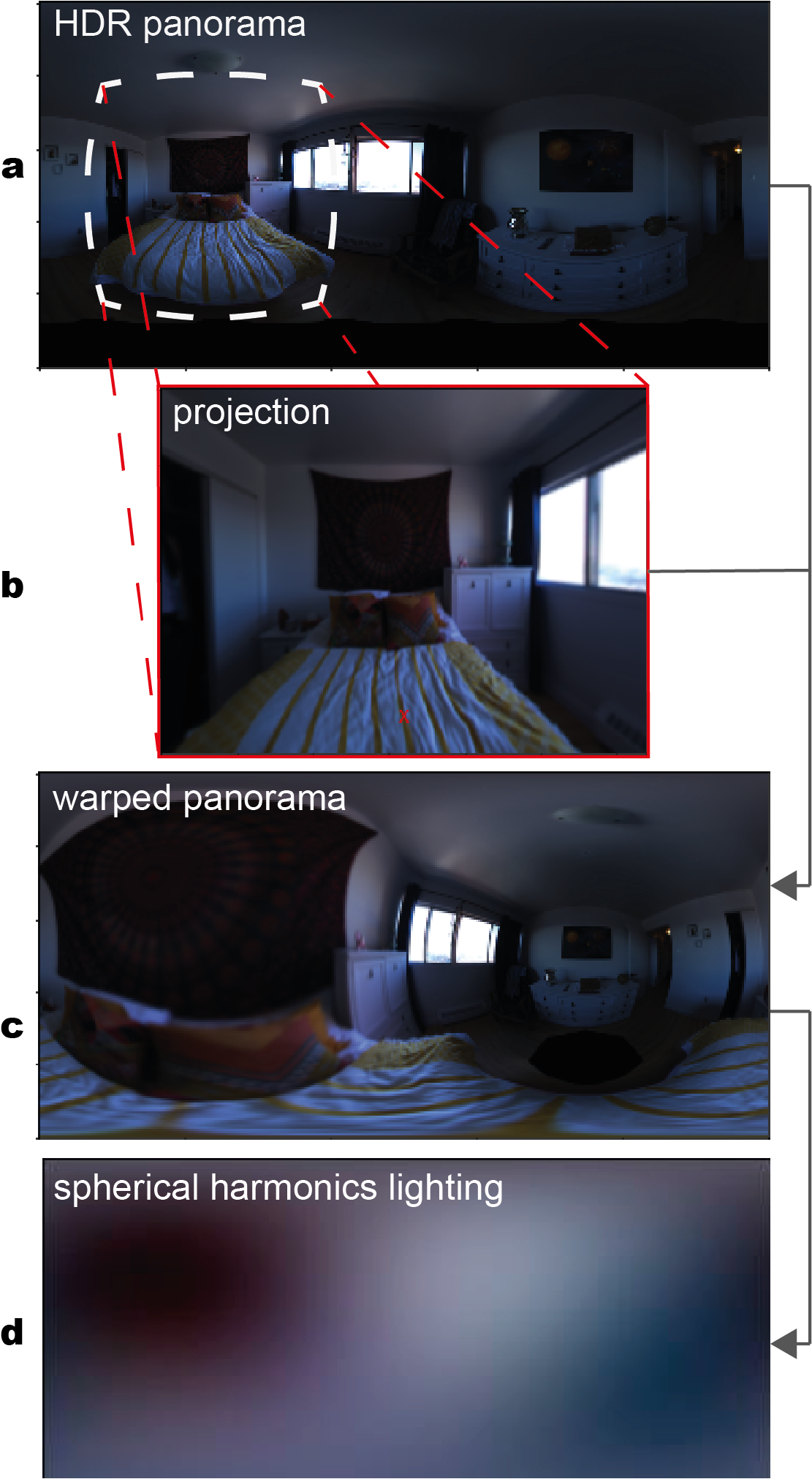}

  \caption{\label{fig:mixed_view}
          Mixed-reality-view processing pipeline. The pipeline takes an HDR panorama (a) and projects this panorama into an LDR mixed-reality-view (b) while generating the HDR spherical harmonics lighting (d) through rotations and a warp operation (c) of the original panorama.}
\end{figure}

The pipeline to obtain a mixed-reality view from the HDR panoramas (Figure \ref{fig:mixed_view}) aims to mimic the behavior of an HMD camera walking through the panorama environment. To accomplish this, we rotate the panorama (Fig. \ref{fig:mixed_view}a) horizontally and vertically, simulating the user's head movements in a XR environment.  We choose a random vertical $\phi$ and horizontal $\theta$ angle in the range of [-15,15] and [-180, 180] degrees, respectively. We simulate a camera by a perspective projection of the environment panorama (Fig. \ref{fig:mixed_view}b). The projection and camera settings were tailored to replicate the camera (HTC Vive integrated camera) employed in our tests. To approximate the spatial changes in the camera position, we use a warp operation $T$ defined by:

\begin{equation}
  T(v, \beta) = \frac{2v_{z} \sin(\beta) + \sqrt{(-2v_{z}\sin(\beta))^2 - 4\norm{v} \sin(\beta) ^2 - 1}}{2 \norm{v} ^2} \\
\end{equation}

where $v$ is the point $(v_{x}, v_{y}, v_{z})$, and $\beta$ is the angle between the camera nadir and the center of projection in the image. This warp operation is based on the warp operation proposed by Gardner et al. \cite{gardner-sigasia-17}. We choose $\beta$ to correspond to the lowest vertical point visible in the mixed-reality-view.

A simple gamma-correction process \cite{reinhard2010high} is applied to projected image (Fig. \ref{fig:mixed_view}b). The resulting LDR image is a mixed-reality-view that corresponds to a rectified crop of the original panorama with a limited field of view that mimics the view in an XR application. 

To obtain the lighting setting of the mixed-reality-view, we project second-order spherical harmonics functions into the (rotated and warped) HDR panorama (Fig. \ref{fig:mixed_view}c). The projection generates nine coefficients, one for each function of the second-order spherical harmonics. Since we intend to estimate the color of the lighting environment, we project a set of $9$ coefficients for each color channel, resulting in $27$ coefficients. Note that all the process to obtain the spherical harmonics lighting is performed in HDR. The $27$ SH coefficients are represented by a lighting map in Figure \ref{fig:mixed_view}d.

 The tuple [LDR mixed-reality-view, HDR SH coefficients] is the sample of our training dataset. To create our dataset, we execute the pipeline eight times for each panorama in the Laval indoor HDR dataset, generating a total of $17152$ samples in our dataset.

\section{Results, Experiments and Performance}\label{sec_results}

In this section, we show the results of our method and discuss the XR applications that are made possible by our lighting estimation method.

\subsection{Lighting Estimation}

\begin{figure}[htb]
  \centering
  \includegraphics[width=.95\linewidth]{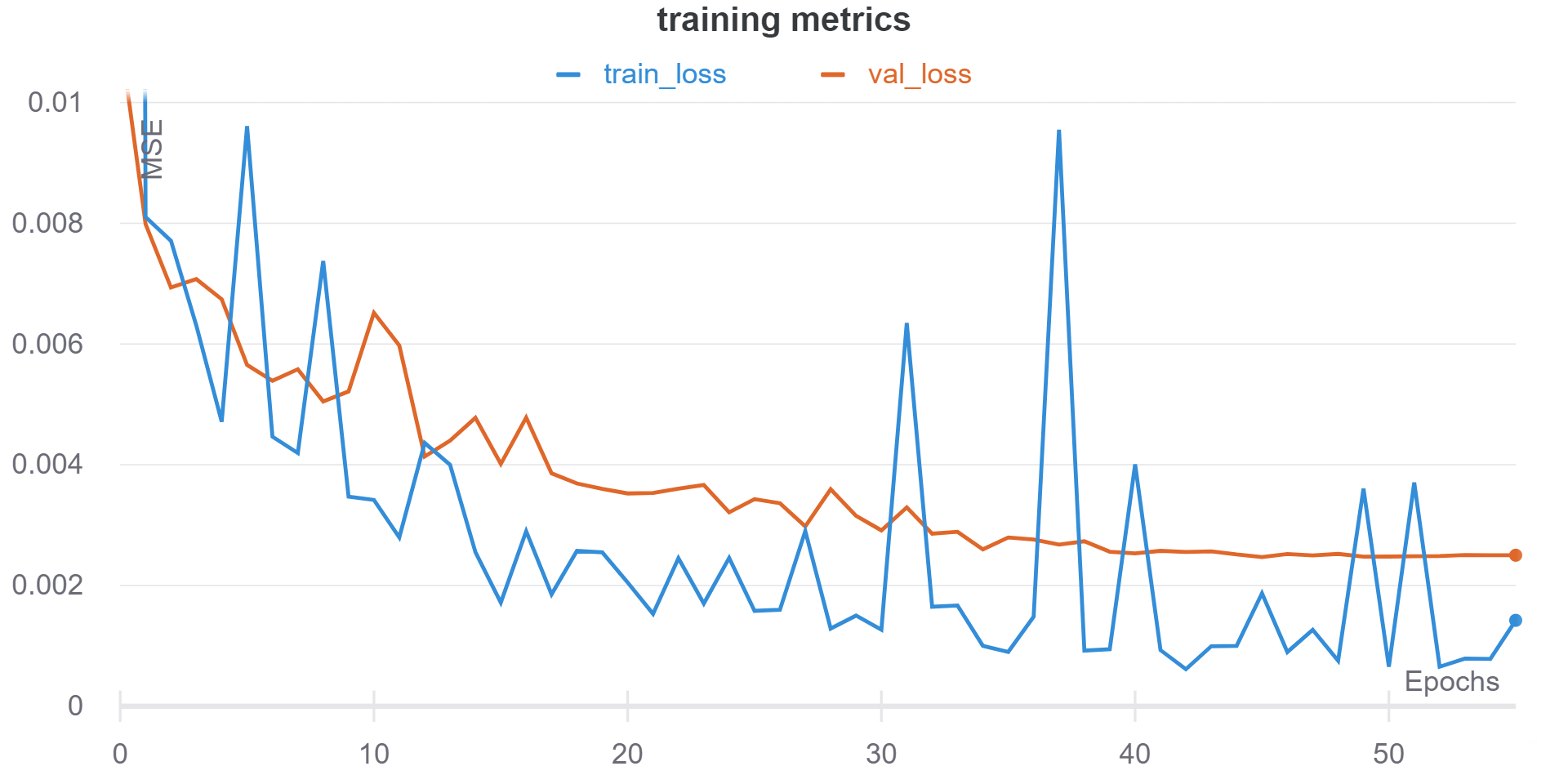}

  \caption{\label{fig:cnn_train}
           Training and validation loss for the lighting estimation model.}
\end{figure}

We split our mixed-reality-view data-set into training, validation, and test sets. We use ~70\% of the data for training ($12008$ samples), 15\% for validation ($2572$ samples), and the remaining 15\% for testing ($2572$ samples). We ensure for our split sets that a mixed-reality-view generated from a specific panorama appears only in a single set; thus, the split produces disjoint sets. Figure \ref{fig:cnn_train} shows the training/validation history of our model. The hyperparameters we use for training: 64 batch size, $10^-4$ learning rate, and Adam optimizer \cite{KingmaAdam} ($b_1=0.9$, $b_2=0.999$). For training purposes, we use an NVIDIA DGX-1 machine, the training process was executed on a P100 GPU, and took 1 hour and 20 minutes. We trained our model for 100 epochs with an early stopping strategy (stop training after 10 epochs without improvements in the validation loss). We achieve convergence of the validation loss after 45 epochs with a mean squared error of 0.002472. 

\begin{table}[htb]
\caption{Mean squared error for the SH coefficients predicted by our model in the test dataset. 25\%, 50\% and 75\% correspond to the 1st quartile, median, and 3rd quartile, respectively.}
\centering
\label{tab:test_errors}
\begin{tabular}{@{}ll@{}}
\\
\toprule
     & MSE    \\ \midrule
mean & \num{7.85e-04}   \\
25\% & \num{1.28e-05}    \\
50\% & \num{5.20e-05}    \\
75\% & \num{3.25e-04}    \\ \bottomrule
\end{tabular}
\end{table}

Table \ref{tab:test_errors} shows the test results of our lighting estimation model. We experimented on images never seen by the neural network during the training phase. We use a mean squared error (MSE) metric to measure the quantitative performance of our model, the SH coefficients are defined in the range [-1, 1]. We obtain the ground truth SH coefficients from the projection of the HDR panoramas on the SH basis. The MSE is calculated by comparing ground truth and predicted SH coefficients. Our model's best predictions (first quartile of test dataset) have an average MSE of \num{1.28e-05}, while our model's worst predictions have an average MSE of \num{3.25e-04}.

\begin{figure}[htb]
  \centering
  \includegraphics[width=\linewidth]{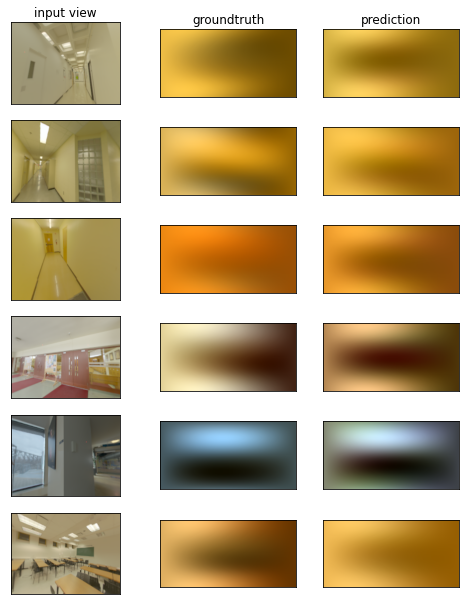}

  \caption{\label{fig:predgt}
           Lighting prediction results. The input for our model (left) is a limited field-of-view image of the scene. The ground truth lighting is obtained from an HDR panorama. The prediction is obtained from the SH coefficients of our model.}
\end{figure}

Figure \ref{fig:predgt} shows examples of the prediction and the ground truth. We represent the SH lighting by irradiance environment maps \cite{Ramamoorthi:2001:ERI}; for visualization purposes, we perform a gamma correction and normalization of the environment maps. The actual difference in rendering using those SH coefficients may be less perceptible. Most of the prediction errors are related to small color differences (such as color saturation) between ground truth and predicted results.

\begin{figure}[h!]
  \centering
  \includegraphics[width=0.95\linewidth]{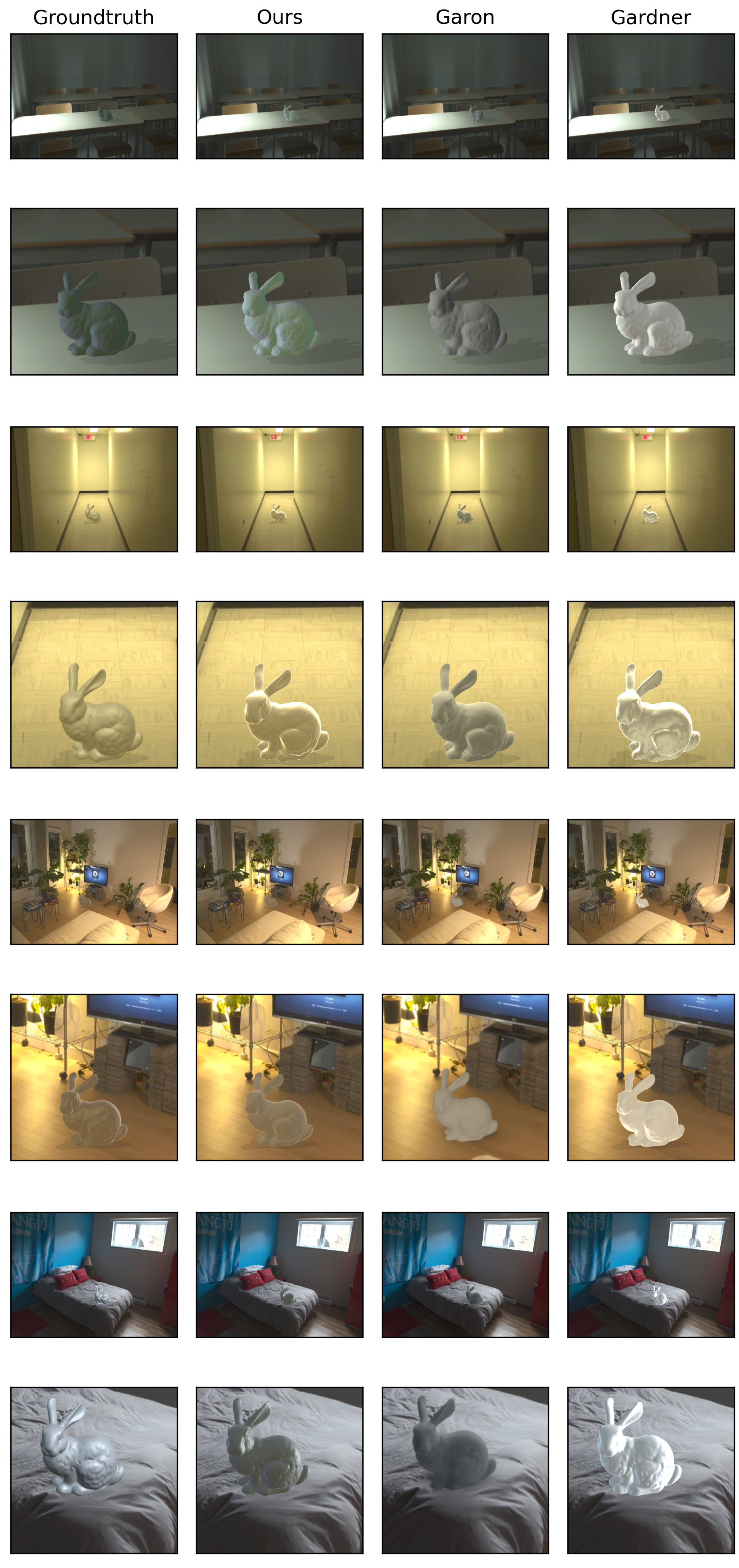}

  \caption{\label{fig:soa_compare}
           Lighting prediction results. The input is a limited field-of-view image of the scene. Our method prediction and Garon et al. \cite{garon2019fast} prediction are SH coefficients of the environment. Gardner et al. \cite{gardner-sigasia-17} prediction is a coarse environment map of the scene.}
\end{figure}

The results presented in Table \ref{tab:soa_errors} and in figures \ref{fig:soa_compare} and \ref{fig:localized} make use of The Laval Indoor Spatially Varying HDR Dataset \cite{Garon_2019_CVPR}. This dataset consists of 20 scenes with HDR panoramas and light probes. The ground truth renderings in our figures and experiments are 3D renderings of a reference object (Stanford bunny) using the original environment map as light probes (we do not approximate the environment map by SH projection) to preserve the lighting information.

The rendering of shadows in the relighting scenes is out of the scope of environment lighting estimation methods. To render the relighting Figures \ref{fig:teaser}, \ref{fig:soa_compare}, and \ref{fig:outdoorRelight}; we project the shadows into a manually placed plane below (and in some cases, behind) the virtual objects. Note that with some geometry information, it is possible to use the environment SH to cast real-time soft shadows \cite{shshadows} for the virtual objects in the scene. Considering that we do not have any geometry information on the real scenes, it is impracticable to generate shadow maps directly. However, some scene registration methods like SLAM \cite{slam} could circumvent this manual tweak in real applications.

Figure \ref{fig:soa_compare} qualitatively compares our model with the state-of-art methods. Our method estimation tends to predict the lighting with more color consistency, blending the virtual object with the environment, while in the other methods, the virtual objects seem artificially placed in the real scenes. Gardner's method usually predicts a mean color environment (closest to grey/brown color), with an overexposed environment light. In comparison, Garon's method seems to produce plausible lights but favoring white light sources. In comparison, Garon's method seems to produce plausible lights but favors white light sources. To quantitatively evaluate and compare our results against the state-of-art methods, we use two metrics, the Structural Similarity (SSIM) error and the Normalized Root Mean-Squared error (NRMSE). To evaluate the color consistency, we convert the RGB images to CIELAB colorspace and compare the $a^*$ and $b^*$ channels of the rendered image using the SSIM error metric. We choose this approach since SSIM is more suitable to measure the perceptual changes in the images, while the chosen ($a^*$, $b^*$) channels represent the chromatic components of the CIELAB colorspace. Table \ref{tab:soa_errors} shows the results of a test that renders lighting estimation for the scenes in the Laval Indoor Spatially Varying HDR dataset. It can be seen that our method produces renderings with a smaller NRMSE and SSIM error than the state-of-art methods.

\begin{table}[]
\caption{Normalized root-mean squared error and Structural similarity error for the renderings using the predicted environment lighting map in the scenes of the Laval Indoor Spatially Varying HDR dataset.}
\centering
\label{tab:soa_errors}
\begin{tabular}{@{}llll@{}}
\\
\toprule
      & Our    & Garon et al. & Gardner et al.\\ \midrule
NRMSE & $0.12 \pm 0.03$ & $0.13 \pm 0.03$ & $0.18 \pm 0.04$  \\
SSIME & $0.12 \pm 0.05$ & $0.14 \pm 0.05$ & $0.15 \pm 0.07$ \\ \bottomrule
\end{tabular}
\end{table}

\subsection{Scenes Relight}

A relighting application consists of changing the illumination of a scene according to a given lighting setting. We test our lighting estimation model in relighting experiments in which stock photos gathered on the internet were used as background, and 3D objects are inserted in the scene. Then, our model's lighting estimation is used to relights the virtual objects and produces the relighted scene.  

We experiment on indoor scenes, where the majority of XR applications happen. As shown in Figure \ref{fig:teaser}, our model is capable of producing plausible lighting for pictures from different sources and lighting settings; this is a fundamental factor since XR applications can significantly differ in respect to the user's environment and device characteristics (camera resolution, field-of-view, lenses distortion).

\begin{figure}[h]
  \centering
  \includegraphics[width=0.9\linewidth]{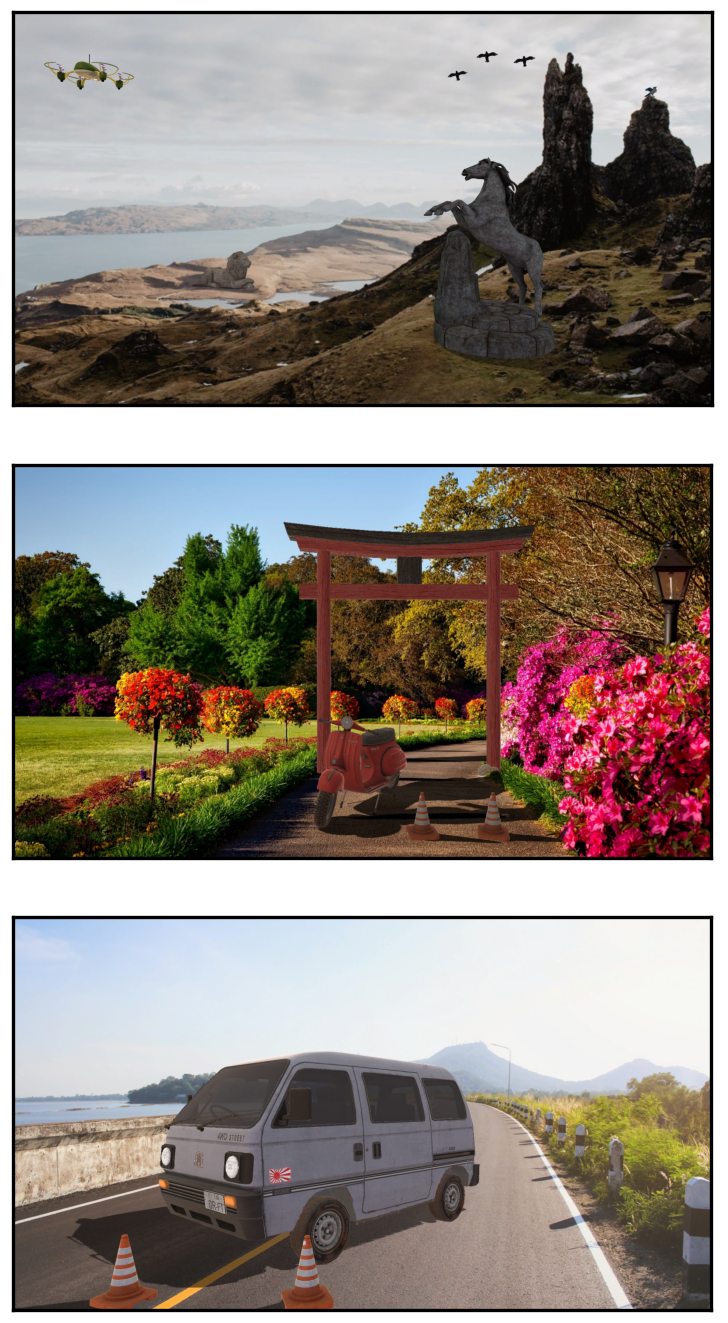}

  \caption{\label{fig:outdoorRelight}
           Relighting of an outdoor scene. Virtual objects are inserted in outdoor stock photos. The virtual objects were relighted with our lighting estimation method.}
\end{figure}

To further demonstrate the generality of our model, we experimented on outdoor scenes (Figure \ref{fig:outdoorRelight}). Outdoor lighting is characterized by high-frequency shadows with a high dynamic range of luminance, dominated by a distant and directional light source (sunlight). For these reasons, it is a challenge to develop a lighting estimation method that works in both outdoor and indoor scenes. Although only indoor images were used in the CNN training, our model estimates the lighting and produces plausible relighting of outdoor scenes. However, some challenging aspects of the outdoor lighting estimation are still are visible in our results. 

We experiment with three outdoor scenarios: an overcast scene, an outdoor scene with a partially visible sky, and a scene with an entire visible sky. In the first row of Figure \ref{fig:outdoorRelight}, our model made a reasonable ambient lighting estimation for this overcast scene. Despite being hard to judge the overall lighting direction in this scene, the intensity of the lighting setting seems plausible. In the middle row of Figure \ref{fig:outdoorRelight}, the lighting estimation intensity seems to be a little bit faded and not prominent as the natural sunlight intensity. In the last row of Figure \ref{fig:outdoorRelight}, the contribution of the sunlight affects the van and the traffic cones differently. This discrepancy is probably caused by the lack of distant illuminants in the indoor scenes that we trained our network. Furthermore, because the sky presents bright spots with intensities similar to the sun, the model can interpret them as light sources, with an equivalent contribution to the scene lighting. To mitigate those limitations, we plan to further investigate outdoor estimation by including HDR outdoor panoramas in our training pipeline and extending our neural network model by estimating parametric sky models, as described in the conclusion section of this paper.

\subsection{Spatially-varying lighting}\label{sec_spatially}

\begin{figure}[htb]
  \centering

  \includegraphics[width=\linewidth]{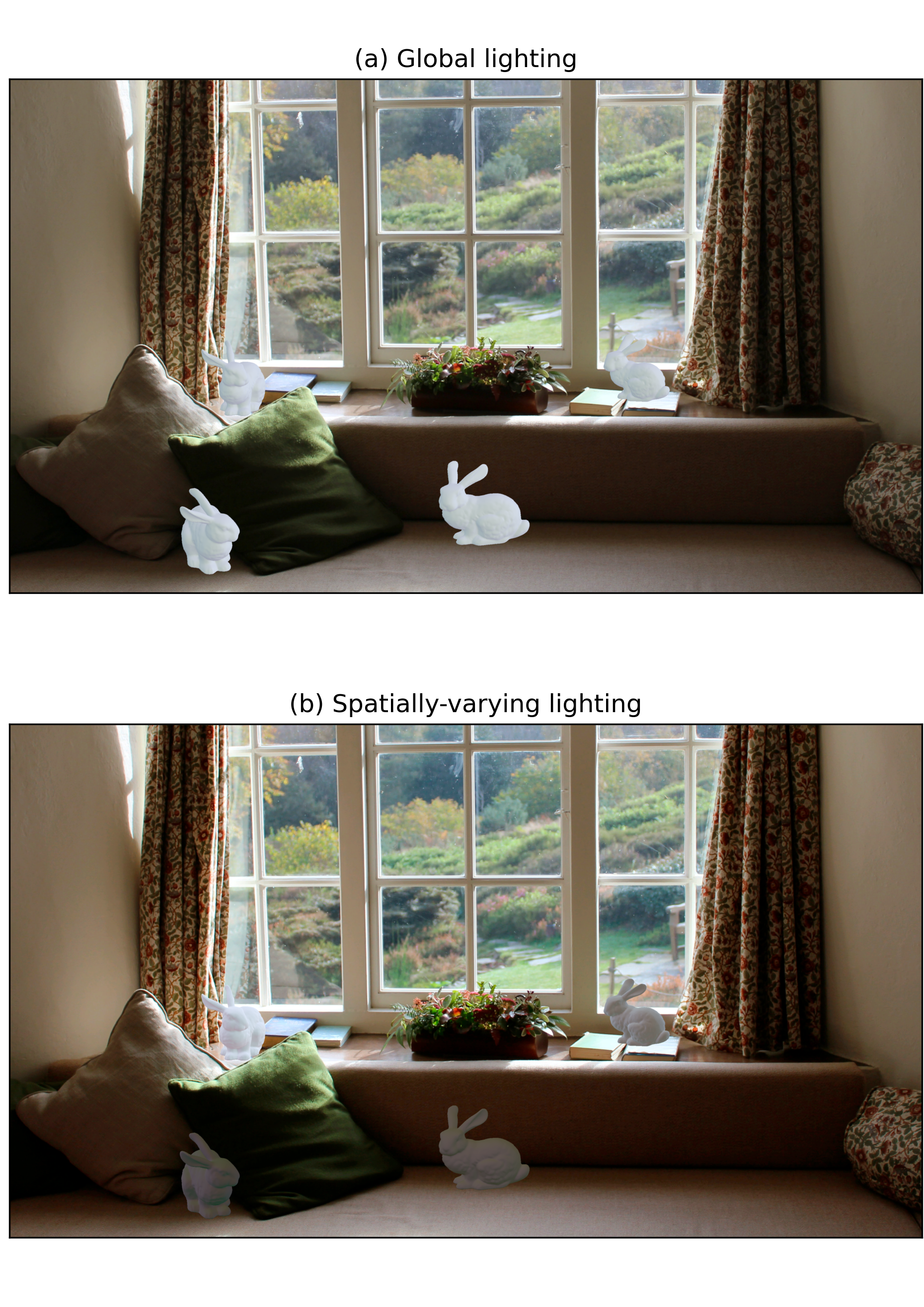}

  \caption{\label{fig:global}
          Global and spatially-varying lighting estimation: using a single image, our method produces a global lighting (a) representing the whole scene. Using a local image for each object (b), our method produces a lighting estimation for each virtual object, resulting in spatially-varying lighting for the scene.}
\end{figure}

Our method is capable of estimating lighting for indoor and outdoor scenes.  However, since the prediction considers the whole scene lighting in a single compact representation, localized lights and occlusion in lighting areas pose a challenging environment for our estimation model. For example, Figure \ref{fig:global}a shows the undesired result when using a single image to estimate the lighting of the entire scene; all 3D virtual objects scattered in the scene are rendered using the same global lighting estimation. 

\begin{figure}[htb]
  \centering

  \includegraphics[width=\linewidth]{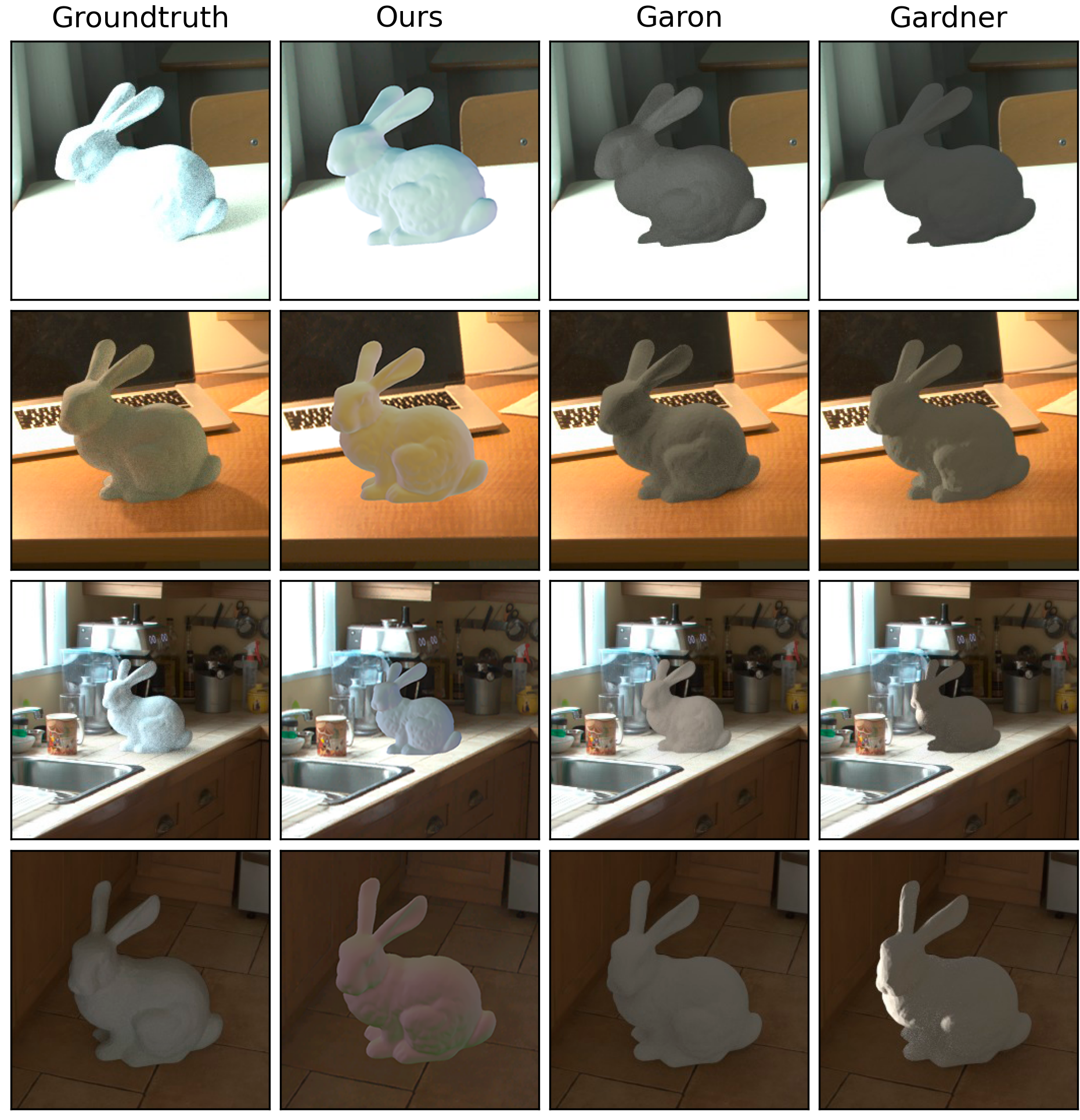}

  \caption{\label{fig:localized}
          Qualitative comparison of state-of-art spatially-varying lighting estimations methods. The virtual bunny is rendered by: ground truth lighting, Our method's estimation, Garon et al.\cite{Garon_2019_CVPR} estimation, and Gardner et al. \cite{gardner-sigasia-17} estimation.}
\end{figure}

We present a spatially-varying lighting estimation approach using our estimation model that circumvents this undesirable global lighting situation, as shown in Figure \ref{fig:global}b. Instead of estimating the lighting of the entire scene, we provide to our lighting estimation model only the localized portion of the scene where the 3D object resides. By using this strategy, our method produces consistent spatially-varying lighting estimation through the entire scene. Figure \ref{fig:localized} shows a qualitative comparison between our model and the state-of-art methods. Our method tends to produce consistent lighting estimations similar to ground truth while maintaining color consistency and convincing light brightness.



\subsection{Performance}
We implemented an application using Unity Engine for rendering and simulation. The application runs on a desktop computer with an AMD Ryzen 7 2700X CPU clocked at 3.9 GHz and 32 GB of RAM and a consumer-grade NVIDIA Geforce 2070 Super GPU. The model inference was implemented using the PyTorch library without any runtime-specific optimization. The performance tests are measured over 1000 inferences of random images scaled to the baseline image resolution of our model ($256 \times 192$ pixels).

As a result, the average inference time for a single image on CPU is $30.7$ milliseconds, while on an NVIDIA Geforce 2070 Super GPU, the model achieves an inference time of $4.4 \pm 0.5$ milliseconds (an average rendering time of 227 frames-per-second), thus satisfying the time budget demanded by XR applications.

We also test how the model performs when providing more than one frame for inference; this use-case is of interest for non-real-time applications, such as offline video processing.  For a batch size of 128 frames, the inference takes $45.9 \pm 7.3$ ms, resulting in an average rendering time of 0.36 ms per frame. Compared to single-frame inference (batch-size = 1), this difference in inference time is expected since CNN's are highly optimized for batched operations.

\begin{figure}[htb]
  \centering

  \includegraphics[width=\linewidth]{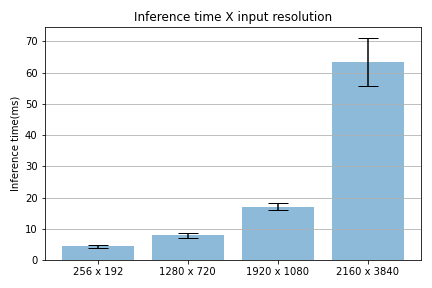}

  \caption{\label{fig:perf_res}
          Inference Time. Measured inference time for input image for various resolutions.}
\end{figure}

Our lighting estimation architecture accepts any image with resolutions higher than 16x16 pixels, meaning that it produces an estimation independent of the input image resolution. However, for optimal estimation results, the input image should be at the exact resolution as the resolution of the samples in the dataset during the CNN's training.

We tested how our model scales concerning the image resolution and inference time. The Figure \ref{fig:perf_res} shows the inference time for the baseline resolution of 256 x 192 px. with $4.4 \pm 0.5$ ms, and other standard resolutions: 1280 x 720 px. with $7.9 \pm 0.8$ ms; 1920x 1080 px. with $17.2 \pm 1.1$ ms, and 2160 x 3840 px. with $63.4 \pm 7.7$ ms. This result shows that the inference-time grows exponentially regarding image resolution. This information allowed us to choose a training resolution for the samples in the dataset that achieves good estimates while maintaining a high inference performance. An additional benefit of this choice is the reduced training time of our model.

It is important to note that the image resolution for the lighting estimation is not necessarily the same as the rendering image; in most applications, the recommended approach of our method is to feed to the estimation model an image scaled to the baseline resolution of 256 x 192 pixels.

Our implementation is limited to second-order spherical harmonics approximations of the environment lighting. One reason for this choice is the diffuse characteristics of the environment lighting. However, it is straightforward to expand our method by changing the size of the last fully connected layer in our neural network to match a higher degree of spherical harmonics approximation. Rendering time is another consideration for using a higher degree of spherical harmonics approximation. Increasing the number of coefficients results in increasing the samples per pixel in the rendering pipeline.


\section{Conclusions}\label{sec_conclusions}

In this work, we introduced a new real-time environment lighting model that is able to compute plausible estimated environment lighting for XR applications directly from mixed-reality-views with no former constraints. Unlike previous approaches, we neither rely on any constraints on the scene geometry and lighting settings nor require the use of probes. 

The environment lighting produced is encoded as $3 \times 9$ spherical harmonic coefficients (9 for each color channel) predicted by a new deep neural network architecture capable of representing the spatially-varying environment lighting with only a single image of the scene.

Our training dataset is defined by a set of mixed-reality-views and SH environment lighting computed from an indoor HDR panorama dataset. We produce a new dataset from the original one by varying user orientation and camera position with respect to each panorama. While camera rotation was simulated by rotating the panoramas horizontally and vertically, the camera spatial variation was obtained by a warping approach. The final dataset consists of an LDR portion that was computed by mapping the HDR images using standard gamma correction and an HDR portion that corresponds to the ground truth SH lighting projected from the HDR panoramas.

The experiments have shown that we can produce plausible environment lighting representations without any strong requirements on the inputs. Compared to some state-of-art works, our work produces a smaller SSIM average error when comparing the predicted environment lighting and the ground truth. We also show that our method can be easily applied to XR and relighting applications.

The product of our model enables XR applications to change and adapt the environment lighting, allowing realistic lighting and immersive simulations. However, some results regarding scenes with many occluded objects can be enhanced. This is still a drawback of our method, which can be explained by the fact that we do not rely on any geometrical information about the scene. We intend to pursue this in future works. Moreover, we believe that the impact of the HDR to LDR mapping on the prediction of the lighting representation must also be more deeply investigated in the future.

Our method is limited to estimating the diffuse component of the environment lighting. Thus the estimation of high-frequency reflection maps is out of the scope of this work. A future direction of this work will be an inclusion of estimations regarding additional light representations along with the SH diffuse lighting. In particular, the estimation of real-time reflection maps for mirror-like surfaces and Spherical Gaussians light \cite{sphericalGaussianTokuyoshi} that better describes all-frequency materials such as highly specular surfaces. Another future extension of our work would be the addition of parametric estimations tailored for outdoor scenes. We envision that incorporating an outdoor discriminator and parametric sky models \cite{hosek2012analytic,  preetham1999practical} into the neural network architecture would make the model robust enough to predict accurate sun lighting for outdoor scenarios while maintaining good performance in indoor scenes.

\section{Acknowledgements}
This research has been supported by the following research agencies: CAPES, CNPq and FAPERJ. We also would like to thanks NVIDIA for providing GPUs and funding this work.

\bibliographystyle{cag-num-names}
\bibliography{biblio}

\end{document}